\documentclass[11pt,a4paper]{article}
\usepackage[hyperref]{eacl2021}
\usepackage{times}
\usepackage{tabularx}
\usepackage{multirow}
\usepackage{latexsym}
\usepackage{booktabs}
\usepackage{graphicx}
\usepackage{xcolor}
\graphicspath{ {./images/} }

\usepackage{microtype}

\usepackage{amssymb}
\usepackage{pifont}

\aclfinalcopy

\title{Predicting Directionality in Causal Relations in Text}

\author{Pedram Hosseini\quad David A. Broniatowski\quad Mona Diab \\
The George Washington University \\
  {\tt \{phosseini,broniatowski,mtdiab\}@gwu.edu}
  }

\date{}

\begin{document}
\maketitle
\begin{abstract}
In this work, we test the performance of two bidirectional transformer-based language models, BERT and SpanBERT, on predicting directionality in causal pairs in the textual content. Our preliminary results show that predicting direction for inter-sentence and implicit causal relations is more challenging. And, SpanBERT performs better than BERT on causal samples with longer span length. We also introduce CREST which is a framework for unifying a collection of scattered datasets of causal relations.

\end{abstract}

\section{Introduction}
\label{sect:intro}
Causal relations play a major role in many tasks in Natural Language Understanding (NLU)~\cite{girju2003automatic} and discourse analysis~\cite{mulder2008understanding,kuperberg2011establishing}. The relation between X and Y is considered causal when X makes Y happen or exist, or vice versa, where X and Y can be an event, state, or object. For example, in "The river had now turned into full \textit{flood} after the \textit{deluge} of rain a few days ago.", \textit{deluge of rain} is the cause of \textit{flood} which happens when there is an overflow of water.

Automatic extraction of causal relations is a challenging task in Natural Language Processing (NLP). Previous work on identifying causal relations is mainly focused on classifying \textit{pairs} as causal/non-causal without necessarily considering \textit{direction} in the pairs. These lines of research either assume a fixed direction between spans in a pair or do not directly test models in predicting the \textit{direction} in a causal pair. For example, \cite{gao2019modeling, ijcai2020-499} focus only on identifying causal pairs in context and not specifying the direction in the pairs, namely which entity caused the other. The knowledge-oriented CNN (K-CNN) model for causal relation extraction model~\cite{li2019knowledge} attempts at finding sentences that include a causal relation only by feeding a pair of spans and a sentence as input to the model. On the other hand, \citep{dunietz2018deepcx} indirectly addresses the direction prediction problem by finding cause and effect spans associated with only explicit causal connectives in context. They adopt a syntactic perspective.

\begin{figure}[h]
\centering
\includegraphics[scale=0.59]{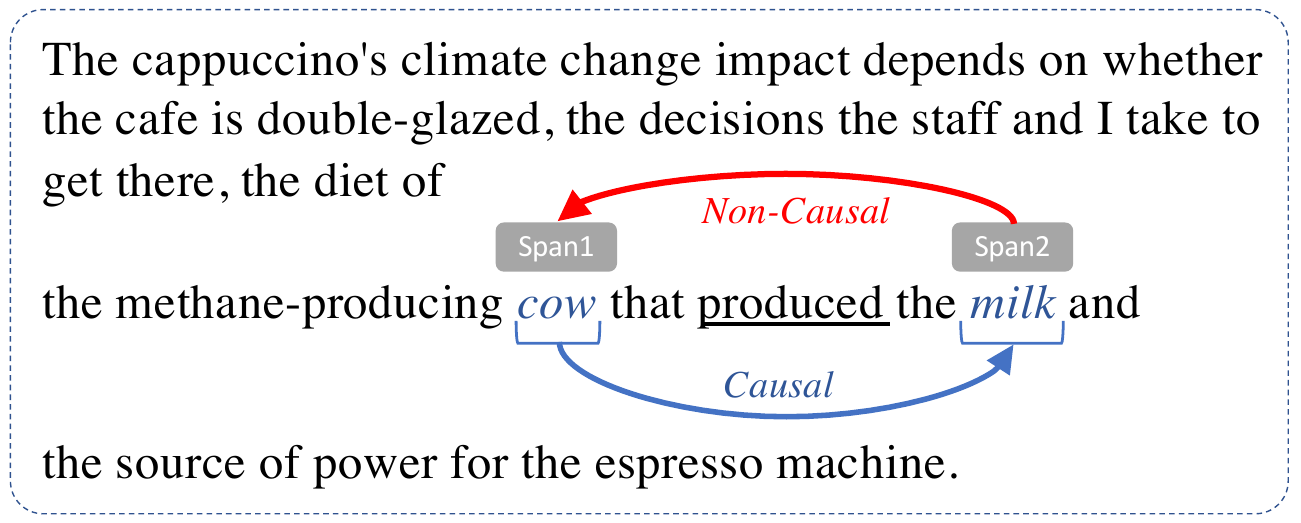}
\caption{\label{fig:example-1}A golden \textit{non-causal} relation that can be \textit{Causal} if we change the direction between spans.}
\end{figure}

\textit{Direction} between spans in a relation may play an important role in the classification of some instances of causal relations. For example, in the non-causal relation shown in Figure~\ref{fig:example-1}, from a commonsense perspective, we know that \textit{milk} cannot produce a \textit{cow} but a \textit{cow} can produce \textit{milk} and be the reason for milk to be produced. Thus, if we treat causal relation classification here simply as a pair classification task with \textit{no direction} provided, we will have a pair that can be marked as both causal and non-causal while the gold label for the pair is non-causal. We then ask a question here: \textbf{RQ)} how well can a model tell the difference between \textit{cause} and \textit{effect} in a causal relation?

In the rest of the paper, first, we discuss the scatteredness of causal relation datasets and introduce a simple framework to unify a collection of well-known datasets of causal relations (\S\ref{sect:data_unification}). Then we explain the preprocessing steps on datasets we chose from this collection for our experiments (\S\ref{sect:data_preprocessing}). We report our experiments on predicting direction in causal relations (\S\ref{sect:experiments}) along with some preliminary results and qualitative error analysis (\S\ref{sect:results}). We conclude this paper with reviewing the related work (\S\ref{sect:related_work}) and discussing next steps (\S\ref{sect:conclusion}).

\section{Unifying Causal Relation datasets}
\label{sect:data_unification}
Data resources of causal relations are rather scattered and do not often follow the same schematic or machine-readable format. Such an incompatibility among resources is not a new problem and it has been slowing down the progress in the classification of semantic relations including causal relations~\cite{hendrickx2010semeval}. That is why we decided to unify a collection of widely used resources of causal/non-causal relations to make them easier for the research community to use. 

In this step, we focused on publicly available datasets in which the \textit{context} associated with a causal relation is available, and we excluded data resources in which a relation is only a pair of words/tokens without any given context such as~\cite{luo2016commonsense}. The main reason is that causal relation/pair classification is mainly a context-dependent task. We also excluded data resources that only contain context and a label (binary or multi-label) without annotating the spans/arguments of a relation such as~\cite{yu2019detecting}. It is worth pointing out that these excluded resources can fit in the format we define and they were just not our focus for now. It is also important to mention that some resources are not publicly available --which itself is another challenge in addition to the incompatibility of schemes-- such as Penn Discourse Treebank (PDTB)~\cite{prasad2008penn}, and some others such as the corpus of Temporal-Causal relations~\cite{bethard2008building} and BECauSE \cite{dunietz-etal-2017-corpus} only share their annotation. \footnote{E.g. not all the annotated samples in~\cite{bethard2008building} and~\cite{dunietz-etal-2017-corpus} can be used unless we have access to Wall Street Journal (WSJ) and The New York Times Annotated corpora.}

In the following subsection, we describe our unifying framework, CREST, in more detail. 

\subsection{CREST: A Causal Relation Schema\footnote{\url{https://github.com/phosseini/CREST}}}
\label{ssect:crest}
CREST is a schema or more precisely, a unifying framework for storing causal/non-causal relations. Various formats and schemes are introduced for annotating causal relations~\cite{bethard2008building,prasad2008penn,o2016richer,mostafazadeh2016caters}. There are two main reasons for the existence of such a wide range of schematics: 1) some data resources, \cite{mirza2014annotating} for instance, are created on top of existing datasets built for tasks other than causal relation extraction so they have to follow the same format as the original dataset. And 2) some more recent datasets include newly defined meta-information for relations that were not annotated in older datasets. For example, Semeval-2007 task 4~\cite{girju-etal-2007-semeval} and Semeval-2010 task 8~\cite{hendrickx2010semeval} data do not have annotations for causal signal or trigger words. But such words are shown to be effective in causal relation extraction~\cite{do2011minimally} and were later annotated in other datasets such as BECauSE~\cite{dunietz-etal-2017-corpus}.

By comparing a wide range of well-known and widely-used datasets of causal relations, we identified the largest common set of features and annotations among them to define CREST. In Table~\ref{tab:crest-fields}, we listed all these features. We call the process of converting relations to CREST as \textit{CRESTing} a dataset.

\begin{table*}[]
\small
\begin{tabularx}{\textwidth}{c|X}
\bottomrule
\multicolumn{1}{c|}{\textbf{(Field name: Field value type)}} & \multicolumn{1}{c}{\textbf{Explanation}} \\ \bottomrule
{\tt original\_id: string}             & Identifier of a relation in the original dataset. \\ \hline
{\tt span1, span2, signal: list}     & These three fields store the arguments of a relation, cause, effect, span (for non-causal relation) and signal/trigger tokens. The value of these fields is a list of token strings. \\ \hline
{\tt context: string}                  & The context in which a relation appears.  \\ \hline
{\tt idx: dictionary}                  & Character indices of \textit{span1}, \textit{span2}, and \textit{signal} tokens in the \textit{context}. Each value is a list of lists where each list in the main list is a pair of start and end indices of a token in context. \\ \hline
{\tt label: integer}                   & Specifies whether there is a causal relation between \textit{span1} and \textit{span2}. 0: non-causal. 1: causal \\ \hline
{\tt direction: integer}                   & Shows the direction of a relation. 0: \textit{span1} $\Longrightarrow$ \textit{span2} and 1: \textit{span1} $\Longleftarrow$ \textit{span2}\\ \hline
{\tt split: integer}                           & Split to which a relation belongs in the source dataset. (0: train, 1: dev, or 2: test set). \\ \bottomrule
\end{tabularx}
\caption{\label{tab:crest-fields}List of features in CREST . We use pandas DataFrame data structure~\cite{mckinney-proc-scipy-2010} to store relations in CREST. All value types are Python values.}
\end{table*}

\section{Data and preprocessing}
\label{sect:data_preprocessing}
We chose two datasets that follow the definition of causal relations from a commonsense perspective to use in our experiments. Compared to other datasets of causal relations such as~\cite{do2011minimally,mirza2014annotating,mostafazadeh2016caters,dunietz-etal-2017-corpus}, these datasets have enough causal relations to allow us to create fairly balanced train and test splits. In the following, we briefly introduce each of them:


\noindent\textbf{Penn Discourse Treebank 3.0 (PDTB3)}~\cite{prasad2008penn} contains annotated samples of discourse connectives, implicit and explicit, and their arguments. These discourse connectives are taken to be the predicates of binary discourse relations including causal relations. There are four coarse-grained discourse relation types at Level-1 including {\tt Contingency} in PDTB3. And Contingency relation itself contains the finer-grained \texttt{Cause} relations at Level-2. We use all Level-3 relations in PDTB3 with Level-2 classes: \textit{Cause}, \textit{Cause+Belief}, and \textit{Cause+SpeechAct}. We excluded the Level-3 \textit{NegResult} relations since in these relations one argument does not cause but \textit{prevents} the effects mentioned in the other argument.

\noindent\textbf{EventStoryLine (ESL) v1.5} is created by crowd-sourcing causal relations between events in news articles~\cite{caselli2018crowdsourcing}. EventStoryline's crowd-sourcing approaches and experiments follow a commonsense reasoning perspective of causality. Causality in EventStoryLine refers to the broader notion of contingent relations rather than a strict causal relation. We extract all \textit{PLOT\_LINK} tags of the two following classes from EventStoryLine: 1) {\tt PRECONDITION}, events which enable or cause another event, or 2) {\tt FALLING\_ACTION} that mark speculations or consequences.

\subsection{Converting relations to sequences}
\label{ssect:relation_conversion}
When we use \textit{relation}, we mean a pair of {\tt span}, in a {\tt context}, where {\tt span} is a sequence of one or more tokens and {\tt context} can be one or more sentences. \textit{Direction} of a relation can be either $span1 \Rightarrow span2$ or $span1 \Leftarrow span2$. In a \textit{causal} relation, the starting span is always \textit{cause} and the ending span is always \textit{effect}. $Span1$ and $span2$ in a relation do not always appear in the same order in {\tt context}.

\begin{figure*}[h]
\centering
\includegraphics[scale=0.62]{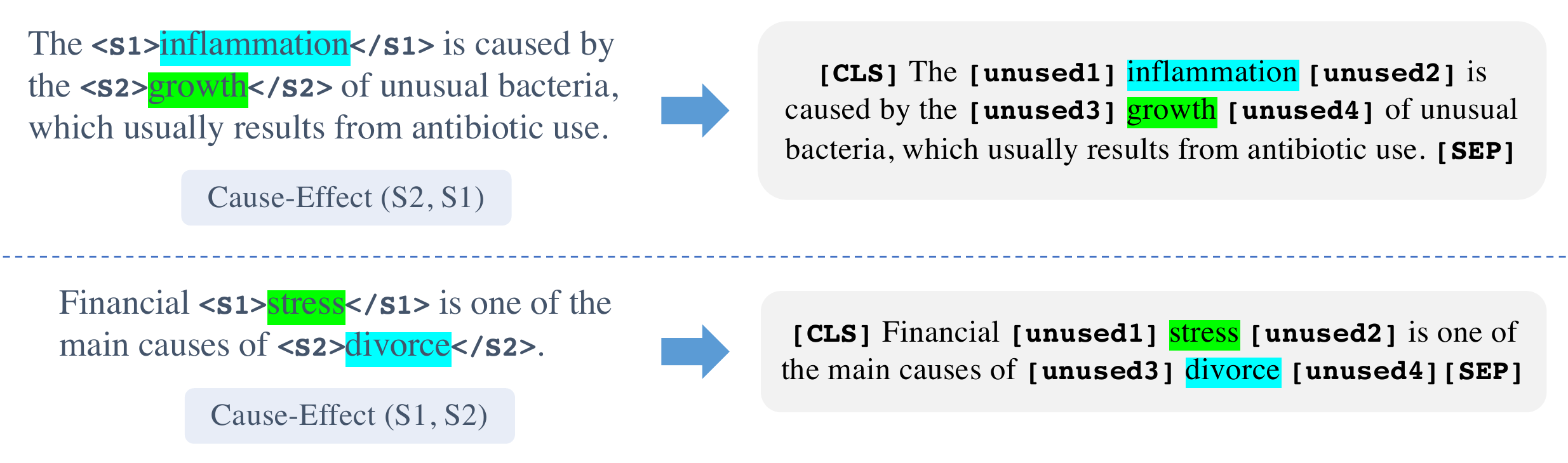}
\caption{\label{fig:convert}Converting an input example to an input sequence with and without \textit{direction}, for the \textit{BertForSequenceClassification} model.}
\end{figure*}

As shown in Figure~\ref{fig:convert}, we convert relations to input sequences. For specifying spans in a sequence, we add the special {\tt[unused*]} tokens from BERT vocabulary\footnote{For models with a different vocabulary than BERT, the {\tt[unused*]} tokens can be replaced by the special tokens in that vocabulary.} to the start and end of each span in the context. It is very important to notice that the order in which {\tt[unused*]} tokens appear in a sequence is always the same, no matter if a relation is \textit{cause-effect} or \textit{effect-cause}. As a result, we do not feed the direction between spans to the models and models will not know the span type or direction at test time. 

\section{Experiments}
\label{sect:experiments}
\noindent{\textbf{Task:}} For a \textit{causal} relation, given a {\tt context}, two {\tt spans} of text, namely {\tt span1} and {\tt span2}, this task is a binary classification of the relation between the two spans into either \textit{cause-effect} or \textit{effect-cause}. The goal in this task is to predict the \textit{direction} in a causal relation.\vspace{1mm}


We chose two transformer-based language representation models including BERT~\cite{devlin2018bert} and SpanBERT~\cite{joshi2020spanbert}, their \textit{base} and \textit{cased} version. BERT is one of the first transformer-based language models that achieved a promising performance across a diverse range of NLP tasks and later inspired the architecture of some other language models~\cite{liu2019roberta,lan2019albert}. SpanBERT is an improved pre-training model, with the same format as BERT, designed to better predict spans of text. In our experiments, we use HuggingFace's \textit{BertForSequenceClassification} which is a BERT model with a sequence classification head on top~\cite{Wolf2019HuggingFacesTS} implemented in PyTorch~\cite{paszke2019pytorch}. In all experiments, we use 4 different random seeds when fine-tuning our models and report the average result.

\subsection{Data splits}
\label{sect:data_splits}
In all our experiments, we divide our data into train, development (dev), and test splits with 80:10:10 ratio. Information about data splits is shown in Table~\ref{tbl:splits}. 

\begin{table}[h]
\centering
\begin{tabular}{c|c|c|c|c}
\bottomrule
\textbf{Data} & \textbf{Source} & \textbf{Train} & \textbf{Dev} & \textbf{Test} \\\bottomrule
$D_1$ & PDTB3 & 5,692 & 727 & 729  \\
$D_2$ & ESL & 1,095 & 169 & 224 \\
\bottomrule
\end{tabular}
\caption{\label{tbl:splits}Number of samples in data splits.}
\end{table}

\noindent{\textbf{Checking context overlaps:}} There is a possibility that causal relations in a dataset share the exact or similar \textit{context}. For example, there are two causal relations, \textit{(cause, effect)} pairs, annotated in the following context: \textit{"A major earthquake struck southern Haiti on Tuesday, knocking down buildings and power lines"}: $("struck", "knocking")$ and $("earthquake", "knocking")$. Even though these relations are annotated in the same context, they are stored as two separate relations. And when splitting the data into train, dev, and test sets, there is a chance that one of these relations be put in train and another one in test set, respectively. When creating splits, we check to make sure there is no such \textit{context} overlap, full or partial between: 1) train+dev and test, and 2) train and dev, since we want our models not to be tested on classifying relations in a context they have already seen and were fine-tuned on.

\section{Results}
\label{sect:results}
Results of experiments are reported in Table~\ref{tbl:task1}. One interesting point to notice is that on PDTB3 where the length of spans is way longer compared to ESL, SpanBERT performs better than BERT. And on ESL where the majority of spans contain one or two tokens, BERT outperforms SpanBERT. Another point to consider is that even though the average performance of best-performing models on \textit{$D_{2}$} is not too low and is higher than random classification (if we consider random as baseline here,) not all models could converge well on \textit{$D_{2}$}. The small size of training data we use for fine-tuning our models on \textit{$D_{2}$} to a certain degree may explain models’ convergence problem. It is worth thinking how we can leverage some recent datasets of causal relations such as GLUCOSE~\cite{mostafazadeh-etal-2020-glucose} to address this problem or focus on methods that require a lesser number of golden annotated samples. Our results here can also be seen as a baseline for future experiments.

\subsection{Qualitative error analysis}
\label{ssect:error_analysis}
For error analysis and for each language model, we looked at all samples misclassified by all models we fine-tuned and tested with different random seeds on each dataset. We observed two main reasons for errors: 1) \textit{inter-sentence} relations in which two spans of a relation appear in two different sentences and far from one another in context, and 2) implicit relations, as we expected. As mentioned earlier, in PDTB3, spans are relatively longer and contain more tokens. It will be interesting to see if we can find a better way to put a model's attention in an input sequence on the main tokens in a long span, for example, by adjusting the start and end tokens for each span ({\tt[unused*]} tokens, in our case) or adding an encoding vector of span boundaries in input sequences to the models.

\begin{table}[ht]
\centering
\begin{tabular}{c|c|c|c|c}
\bottomrule
\textbf{Data} & \textbf{Model} & \textbf{P} & \textbf{R} & \textbf{F1} \\ \bottomrule
\multirow{3}{*}{$D_{1}$} & Random & 0.54 & 0.5 & 0.52 \\
& BERT & 0.83 & 0.82 & 0.83 \\
& SpanBERT & 0.86 & 0.85 & \textbf{0.86} \\ \hline
\multirow{3}{*}{$D_{2}$} & Random & 0.5 & 0.47 & 0.48 \\
& BERT & 0.8 & 0.78 & \textbf{0.79} \\
& SpanBERT & 0.76 & 0.67 & 0.71 \\
\bottomrule
\end{tabular}
\caption{\label{tbl:task1}Precision, Recall, and F1 of evaluating BERT and SpanBERT models on predicting direction in a causal relation.}
\end{table}

\section{Related work}
\label{sect:related_work}
Work on \textit{automatic} extraction of causal relations started around the late 80s and early 90s with a focus on rule-based methods with a substantial amount of manual work~\cite{joskowicz1989deep,kaplan1991knowledge,garcia1997coatis,khoo1998automatic}. Starting 2000, a combination of learning-based and rule-based methods were employed to improve the quality of automatic causality extraction in text~\cite{girju2003automatic,chang2004causal,chang2006incremental,blanco2008causal,do2011minimally,hashimoto2012excitatory,hidey-mckeown-2016-identifying}. More recently, word-embedding and language representation models started to emerge in work around causal relation classification~\cite{dunietz2018deepcx,pennington2014glove,dasgupta2018automatic,gao-etal-2019-modeling}. 

To the best of our knowledge, the closest study to ours is~\cite{bhagat2007ledir} where LEDIR is introduced as a method to predict directionality in inference rules. 

\section{Conclusion}
\label{sect:conclusion}
In this work, we evaluated the performance of two bidirectional transformer-based language models on predicting the direction in causal relations in text. Our preliminary results show that finding directionality of inter-sentence and implicit causal pairs is more challenging and SpanBERT performs better than BERT on classifying causal relations with longer span length. We also introduced a framework, CREST, for unifying a collection of scattered causal relation datasets. As our next steps, we will work on different methods of feeding input sequences to our causal relation classification models such that boundaries of spans in context are less latent. And, we will continue unifying new datasets of causal relations and add counterfactual relations to our unified collection as well.


\bibliography{anthology,eacl2021}
\bibliographystyle{acl_natbib}

\appendix

\section{Experiments}
\begin{figure}[h]
\centering
\includegraphics[scale=0.4]{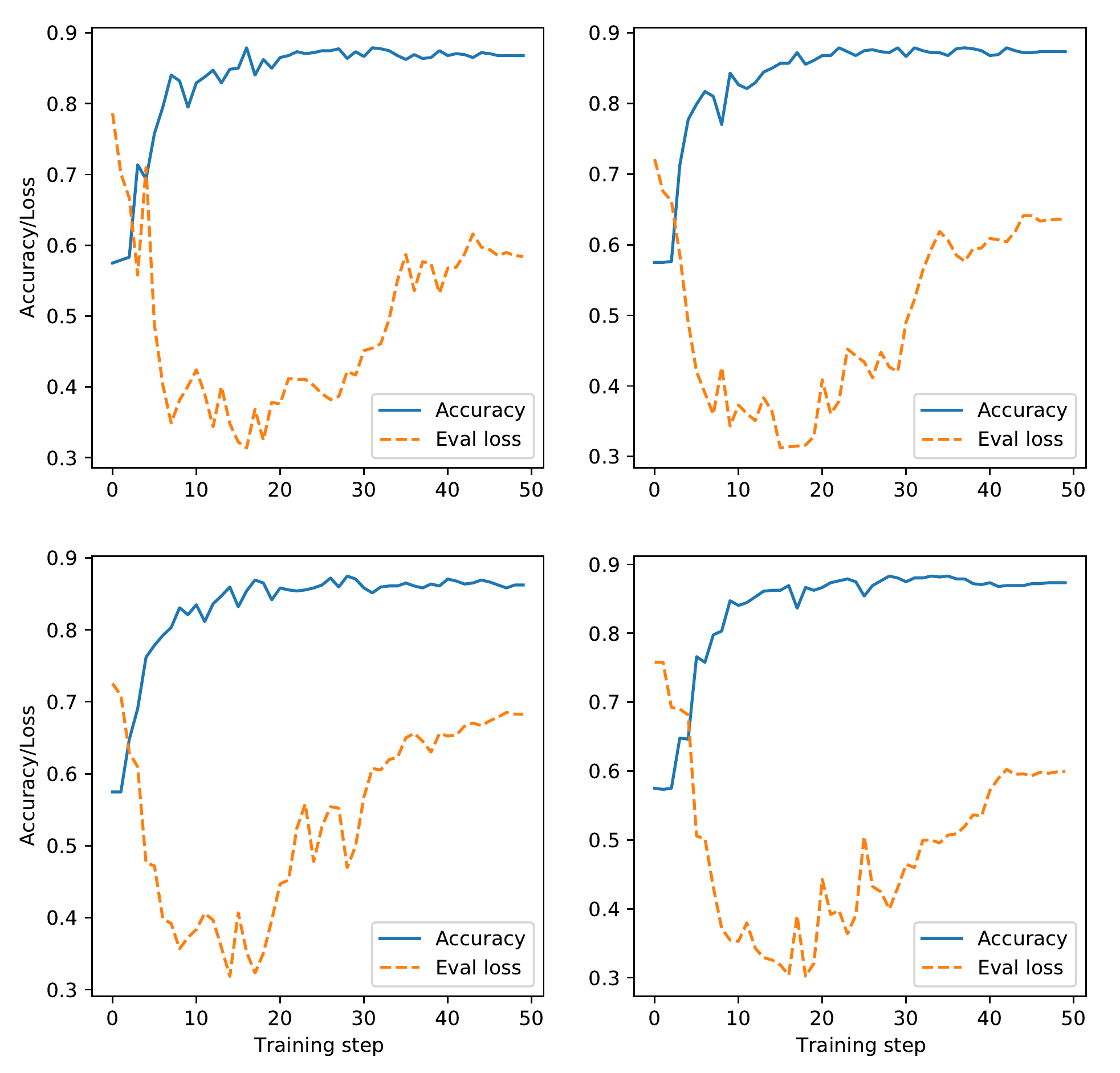}
\caption{\label{fig:b_plot_9}Accuracy and evaluation loss of BERT model on $D_{1}$'s development set.}
\end{figure}

\begin{figure}[h]
\centering
\includegraphics[scale=0.4]{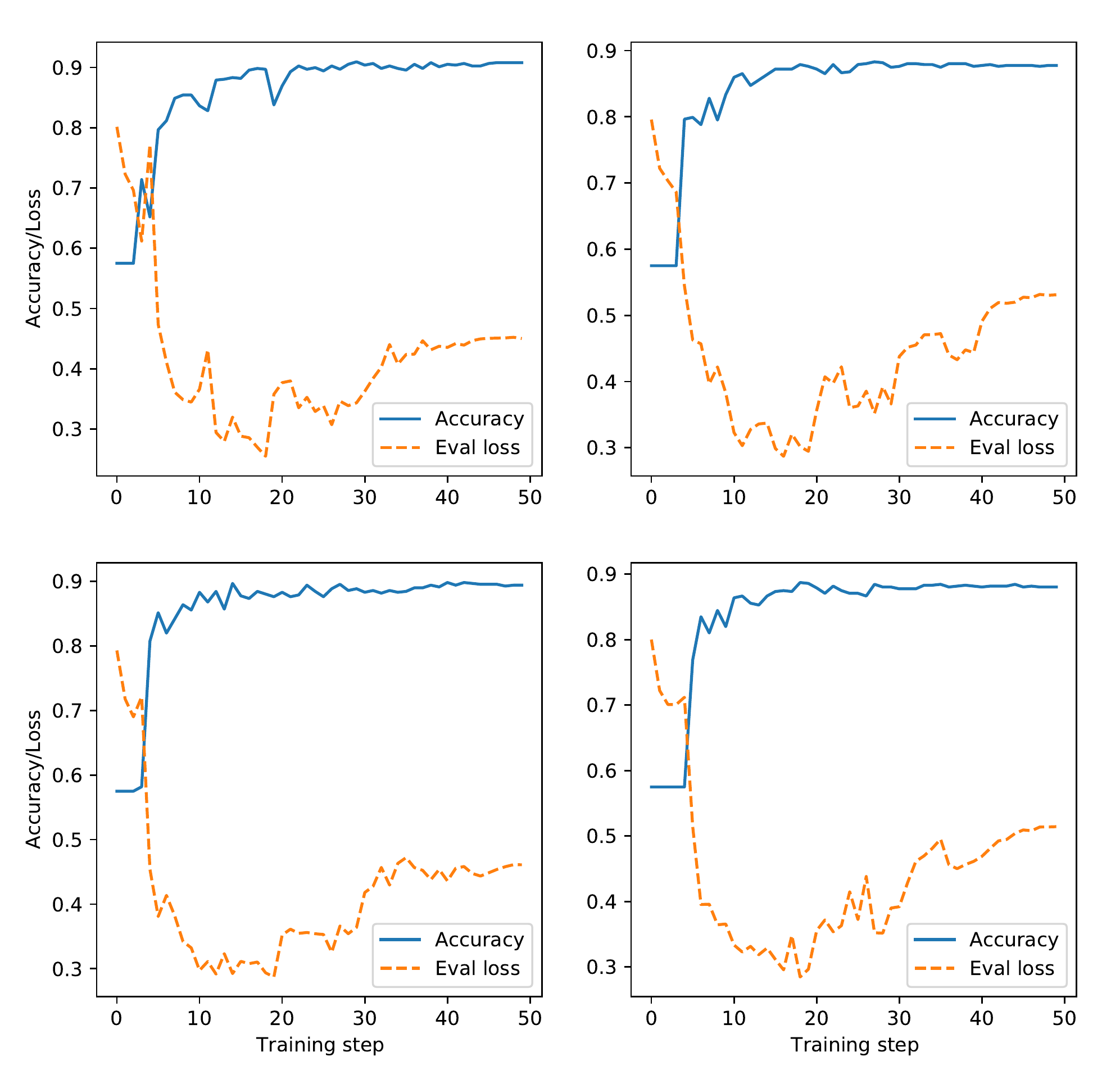}
\caption{\label{fig:sb_plot_9}Accuracy and evaluation loss of SpanBERT model on $D_{1}$'s development set.}
\end{figure}

\begin{figure}[h]
\centering
\includegraphics[scale=0.4]{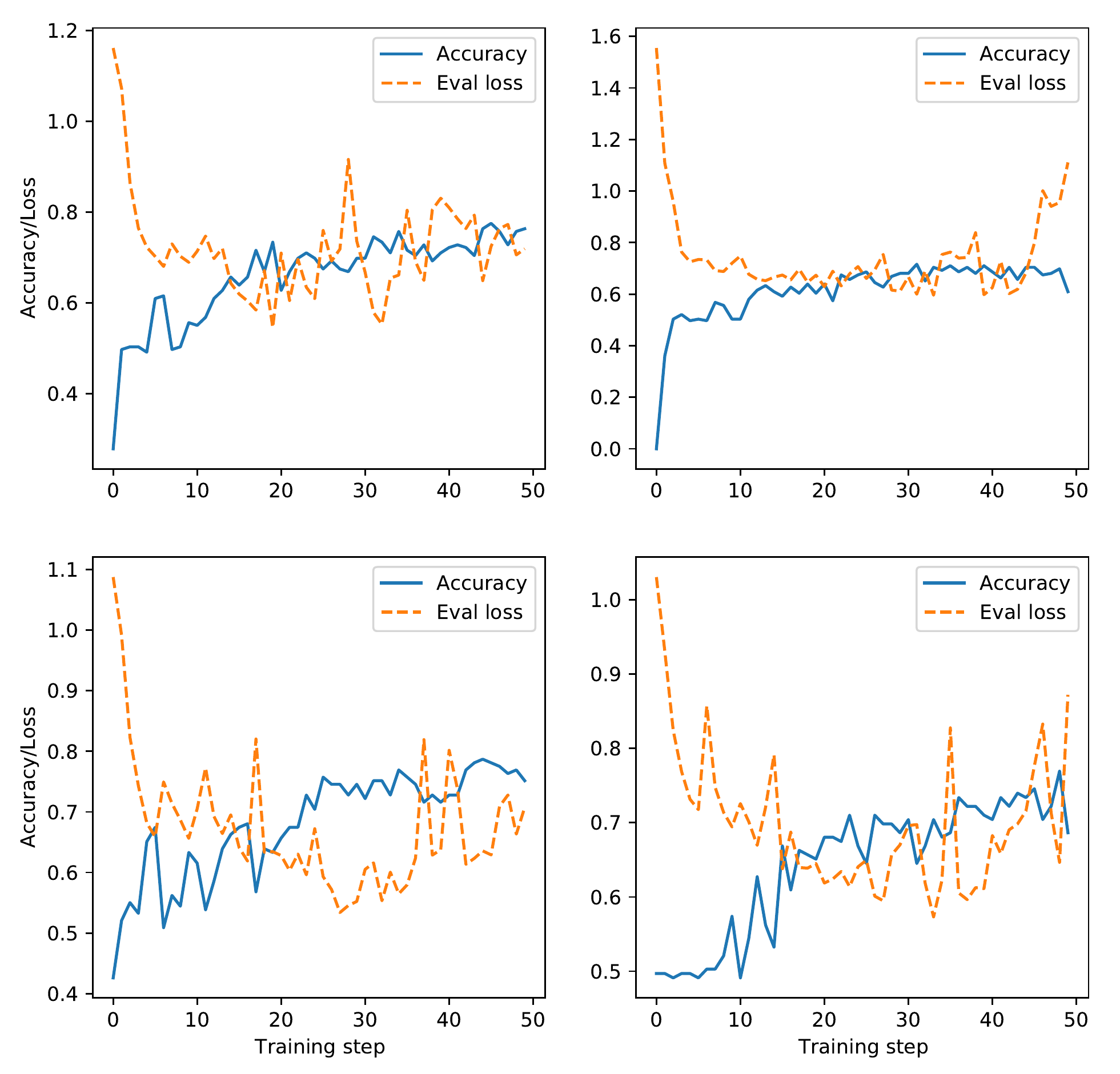}
\caption{\label{fig:b_plot_5}Accuracy and evaluation loss of BERT model on $D_{2}$'s development set.}
\end{figure}

\begin{figure}[h]
\centering
\includegraphics[scale=0.4]{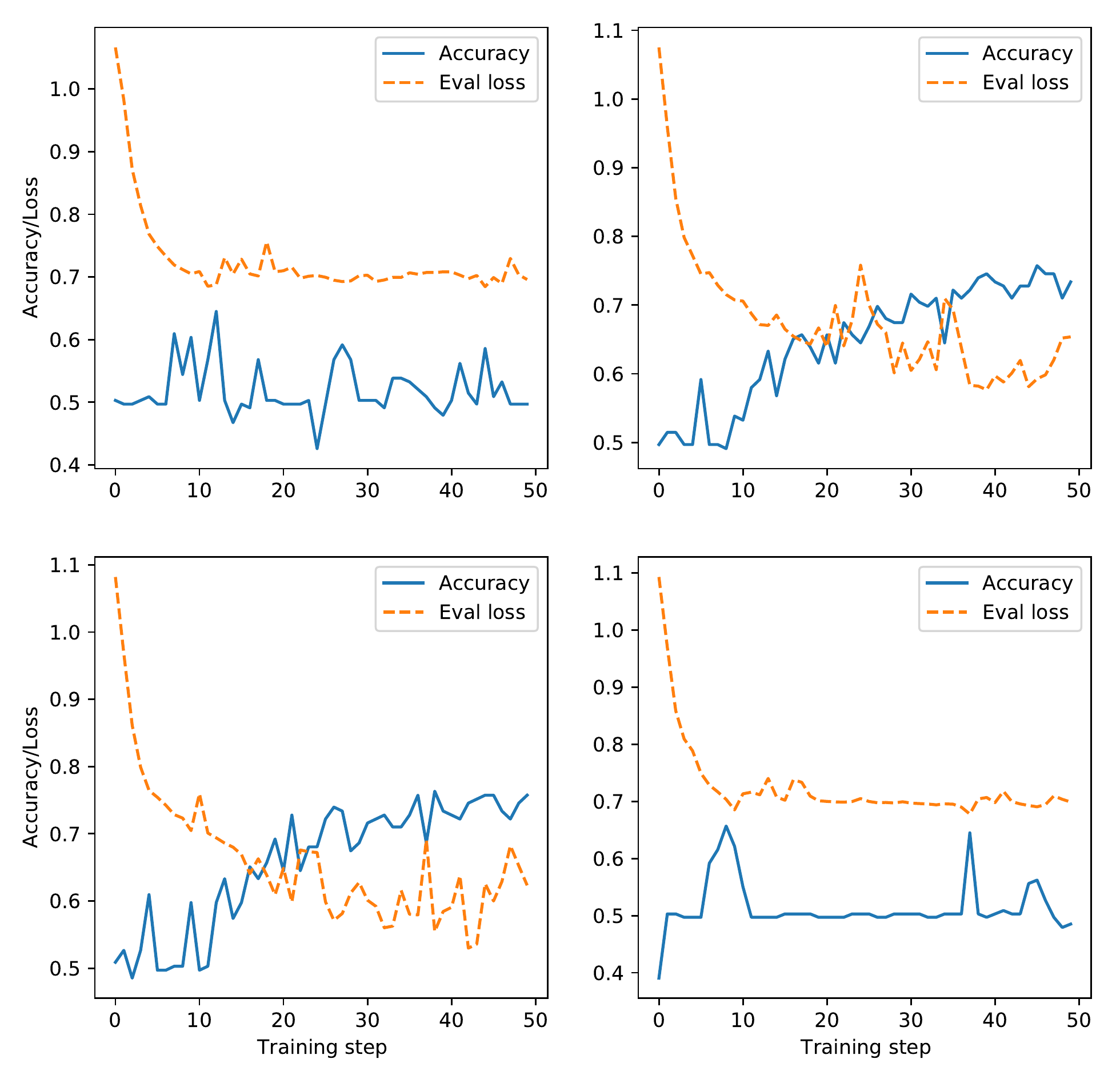}
\caption{\label{fig:sb_plot_5}Accuracy and evaluation loss of SpanBERT model on $D_{2}$'s development set.}
\end{figure}

Accuracy and evaluation loss plots for our experiments on dev set are shown in Figures~\ref{fig:b_plot_9}, \ref{fig:sb_plot_9}, \ref{fig:b_plot_5}, and~\ref{fig:sb_plot_5}. As can be seen in Figures~\ref{fig:b_plot_5}, \ref{fig:sb_plot_5}, models fine-tuned on $D_{2}$ (EventStoryLine v1.5) could not converge well in the majority of cases which to a certain degree can be explained by the small set of causal relations we use for fine-tuning our models.

We fine-tuned our models with sequence length of 128 and the following hyperparameter values: Batch size: 16, learning rate: 2e-5, and number of epochs: 10. All experiments were run on an Amazon AWS \textit{p3.2xlarge} EC2 instance with one Tesla V100 GPU.

\section{CRESTed Datasets}
In this section, we briefly introduce datasets we have aggregated and CRESTed so far. Some statistics related to these datasets are shown in Table~\ref{tbl:dataset-features}.

\begin{table}[]
\centering
\begin{tabular}{c|c|c|c}
\bottomrule
\textbf{ID} & \textbf{Dataset} & \textbf{Signal} &\textbf{Causal} \\ \bottomrule
1 & Semeval-2007 task 4 & \ding{55} &  114             \\
2 & Semeval-2010  task  8 & \ding{55} &  1,331              \\
3 & EventCausality  & \ding{55}  &  485              \\
4 & Causal-TimeBank & \ding{51}  &  318       \\
 5 & EventStoryLine v1.5 & \ding{51} &  2,608 \\
6 & CaTeRS & \ding{55} &  308 \\
7 & BECauSE v2.1 &\ding{51} &  554 \\
8 & COPA & \ding{55} & 1,000 \\
9 & PDTB3 & \ding{51} & 7,991 \\
\bottomrule
\end{tabular}
\caption{\label{tbl:dataset-features}List of CRESTed datasets. \textbf{Signal} refers to signal words/tokens or markers annotated for a relation.}
\end{table}

\noindent\textbf{Semeval-2007 task 4} contains samples of different semantic relations between nominals in text~\cite{girju-etal-2007-semeval}. Cause-Effect is one of these semantic relations. Nominal in this dataset is defined as a noun or base noun phrase excluding named entities. Samples in Semeval-2007 task 4 are initially extracted based on predefined patterns and then manually annotated for the final label. \textbf{Semeval-2010 task 8}~\cite{hendrickx2010semeval} is very similar to the Semeval-2007 task 4 dataset and mainly follows the same schema. The main goal in creating Semeval-2010 task 8 was to have a standard benchmark dataset for multi-way semantic relation classification in context.

\noindent\textbf{EventCausality} is a dataset created for evaluation purposes using news articles~\cite{do2011minimally}. There are two types of relations annotated in this dataset including \textit{causality} and \textit{relatedness}\footnote{Since the existence of causality in \textit{Relatedness} relations is debatable, we excluded these relations when converting relations from EventCausality.}, C and R, respectively. Causality relations in EventCausality are manually annotated for pairs of events based on two rules: 1) Cause event should temporally precede the Effect event and 2) Effect event occurs because the Cause event occurs.

\noindent\textbf{Choice of Plausible Alternatives (COPA)} is a tool for evaluating models' performance in commonsense causal reasoning~\cite{roemmele2011choice}. COPA consists of 1000 questions, split equally into development and test sets. Each question is composed of a premise of and two alternatives where premise is either cause/effect and one of the alternatives more plausibly has a causal relation with the premise. When CRESTing COPA, we also store pairs of premises and their less plausible alternatives. 

\noindent\textbf{Causal-TimeBank} contains explicit causal relations between event pairs in a sentence~\cite{mirza2014annotating}. For annotating causal relations, C-LINKs, expressions containing affect verbs, link verbs, causative conjunctions, and prepositions are used in addition to basic construction for CAUSE, ENABLE, and PREVENT categories of causation and periphrastic causatives. In Causal-TimeBank, polarity, factuality, and certainty are annotated for events involved in a causal relation. Signals of causality for CLINKs are annotated in Causal-TimeBank as well.

\noindent\textbf{CaTeRS} has annotations of explicit and implicit causal relations between events from a collection of stories~\cite{mostafazadeh2016caters}. In CaTeRS, causal relations between events are annotated more from a commonsense reasoning perspective than based on the presence of causal markers. Moreover, annotated relations are either in a sentence or across sentences in a story which makes CaTeRS a proper dataset especially for evaluating the performance of models on identifying inter-sentence causal relations.

\noindent\textbf{EventStoryLine v1.5} EventStoryLine is created by crowd-sourcing causal relations between events in news articles~\cite{caselli2018crowdsourcing}. EventStoryline's crowd-sourcing approaches and experiments follow a commonsense reasoning perspective of causality, the approach adopted by CaTeRs as well. Causality in EventStoryLine refers to the broader notion of contingent relations rather than a strict causal relation. When converting EventStoryLine relations to CREST, we extract all \textit{PLOT\_LINK} tags of the two following classes: 1) {\tt PRECONDITION}, events which enable or cause another event, or 2) {\tt FALLING\_ACTION} that mark speculations or consequences. 

\noindent\textbf{BECauSE} is a bank of effects and causes that are explicitly stated in the context. BECauSE uses a variety of constructions to express causal relations. Causal relations in this dataset are not necessarily relations that hold from a philosophical perspective in the real-world but relations that are expressed in context as causal~\cite{dunietz-etal-2017-corpus}. Every relation in BECauSE has a signal word associated with it.

\noindent\textbf{Penn Discourse Treebank (PDTB3)}~\cite{prasad2008penn} contains annotated samples of discourse connectives, implicit and explicit, and their arguments. These discourse connectives are taken to be the predicates of binary discourse relations including causal relations. There are four coarse-grained discourse relation types at Level-1 including {\tt Contingency} in PDTB3. And Contingency relation itself contains the finer-grained \texttt{Cause} relations at Level-2. We use all Level-3 relations in PDTB3 with Level-2 classes: \textit{Cause}, \textit{Cause+Belief}, and \textit{Cause+SpeechAct}. We excluded the Level-3 \textit{NegResult} relations since in these relations one argument does not cause but \textit{prevents} the effects mentioned in the other argument.

\end{document}